\documentclass{scrartcl}
\usepackage[T1]{fontenc}
\usepackage[utf8]{inputenc}
\usepackage{amsmath, amsthm, amssymb}
\usepackage{dsfont}
\usepackage{graphicx}
\usepackage[format=plain]{caption}
\usepackage{capt-of}
\usepackage{subcaption}
\usepackage{bm}
\usepackage{tikz}
\usetikzlibrary{arrows,shapes,trees,calc,through}

\renewcommand{\phi}{\varphi}

\numberwithin{equation}{section}

\newtheorem{thm}{Theorem}

\theoremstyle{definition} \newtheorem{algo}[thm]{Algorithm}
\theoremstyle{definition} 
\theoremstyle{definition}

\begin{document}

\title{Compensating data shortages in manufacturing with monotonicity knowledge}

\author{Martin von Kurnatowski$^{1}$, Jochen Schmid$^{1}$, Patrick Link$^{2}$,\\ Rebekka Zache$^{2}$, Lukas Morand$^{3}$, Torsten Kraft$^{3}$,\\ Ingo Schmidt$^{3}$, Anke Stoll$^{2}$ \\  
\small $^1$ Fraunhofer Institute for Industrial Mathematics (ITWM), 67663 Kaiserslautern, Germany\\ 
\small $^2$ Fraunhofer Institute for Machine Tools and Forming Technology (IWU), 09126~Chemnitz,\\ \small Germany\\
\small $^3$ Fraunhofer Institute for Mechanics of Materials (IWM), 79108 Freiburg, Germany\\
\small martin.von.kurnatowski@itwm.fraunhofer.de}  

\date{}


\maketitle

\begin{abstract}
Optimization in engineering requires appropriate models. In this article, a regression method for enhancing the predictive power of a model by exploiting expert knowledge in the form of shape constraints, or more specifically, monotonicity constraints, is presented. Incorporating such information is particularly useful when the available data sets are small or do not cover the entire input space, as is often the case in manufacturing applications. The regression subject to the considered monotonicity constraints is set up as a semi-infinite optimization problem, and an adaptive solution algorithm is proposed. The method is applicable in multiple dimensions and can be extended to more general shape constraints. It is tested and validated on two real-world manufacturing processes, namely laser glass bending and press hardening of sheet metal. It is found that the resulting models both comply well with the expert's monotonicity knowledge and predict the training data accurately. The suggested approach leads to lower root-mean-squared errors than comparative methods from the literature for the sparse data sets considered in this work.
\end{abstract}


\section{Introduction} \label{sec:intro}

Conventional machine learning models are purely data-based. Accordingly, the predictive power of such models is generally bad if the underlying training data $\mathcal{D} = \{(\bm{x}_l,t_l): l \in \{1,\dots,N\} \}$ is insufficient. Unfortunately, such data insufficiencies occur quite often, and they can come in one of the following forms: On the one hand, the available data sets can be too small and have too little variance in the input data points $\bm{x}_1, \dots, \bm{x}_N$. This problem frequently occurs in manufacturing~\cite{Weichert.2019} because varying the process parameters beyond well tested operating windows is usually costly. On the other hand, the output data $t_1, \dots, t_N$ can be too noisy. 

Aside from potentially insufficient data, however, one often also has additional knowledge about the relation between the input variables and the responses to be learned. Such extra knowledge about the considered process is referred to as expert knowledge in the following. In  \cite{MacInnes.2010} the interaction of users with a software is tracked  to capture their expert knowledge in a general form as training data for a classification problem. In \cite{Heese.2019} expert knowledge is used  in the form of a specific algebraic relation between in- and output to solve a parameter estimation problem with artificial neural networks. Such informed machine learning~\cite{Rueden.2019} techniques beneficially combine expert knowledge and data to build hybrid or gray-box models~\cite{Johansen.1996, Mangasarian.2007, Mangasarian.2008, Cozad.2015, Wilson.2017, Wilson.2019, Asprion.2019, Heese.2020}, which predict the responses more accurately than purely data-based models. In other words, by using informed machine learning techniques, one can compensate data insufficiencies with expert knowledge. 

An important and common type of expert knowledge is prior information about the monotonicity behaviour of the unknown functional relationship $\bm{x} \mapsto y(\bm{x})$ to be learned. A lot of concrete application examples with monotonicity knowledge can be found in~\cite[Sec.~4.1]{Altendorf.2005} or~\cite[Sec.~1]{Kotowski.2009}, for instance. The present article exclusively deals with regression under such monotonicity requirements. For classification under monotonicity constraints, see e.g.~\cite{Kotowski.2009, Lauer.2007}. Along with convexity constraints, monotonicity constraints are probably the most intensively studied shape constraints~\cite{Groeneboom.2014} in the literature and correspondingly, there exist  plenty of different approaches to incorporate monotonicity knowledge in a machine learning model. See~\cite{Gupta.2016} for an extensive overview. Very roughly, these approaches can be categorized according to when the monotonicity knowledge is taken into account: in or only after the training phase. In the terminology of~\cite{Rueden.2019}, this corresponds to the distinction between knowledge integration in the learning algorithm or in the final hypothesis, respectively. 

A lot of methods -- especially from the mathematical statistics literature -- like~\cite{Mukerjee.1988, Mammen.1991, Mammen.2001, Hall.2001, Dette.2006, Dette.2006b, Lin.2014} incorporate monotonicity knowledge only after training. These articles start with a purely data-based initial model, which in general does not satisfy the monotonicity requirements, and then monotonize this initial model according to a suitable  monotonization procedure like projection~\cite{Mukerjee.1988, Mammen.1991, Mammen.2001, Lin.2014}, rearrangement~\cite{Dette.2006, Dette.2006b, Chernozhukov.2009} or tilting~\cite{Hall.2001}. Among other things, it is shown in the mentioned articles that, in spite of noise in the output data, the arising monotonized models are close to the true relationship for sufficiently large training data sets. Summarizing, these articles show that for large data sets noise in the output data can be compensated by monotonization to a certain extent. 

In contrast to that, in some works like~\cite{Altendorf.2005, Lauer.2008, Chuang.2020, Riihimaki.2010, Neumann.2013, Gupta.2016} monotonicity knowledge is incorporated already in training. In these articles, the monotonicity requirements are added as constraints -- either hard~\cite{Gupta.2016, Lauer.2008, Riihimaki.2010, Neumann.2013} or soft~\cite{Altendorf.2005, Lauer.2008} -- to the data-based optimization of the model parameters. In~\cite{Riihimaki.2010} and~\cite{Altendorf.2005}, probabilistic monotonicity notions are used. In~\cite{Lauer.2008, Chuang.2020, Riihimaki.2010, Neumann.2013} support vector regressors in the linear-programming or the more standard quadratic-programming form, Gaussian process regressors, and neural network models are considered, respectively, and monotonicity of these models is enforced by constraints on the model derivatives at predefined sampling points~\cite{Lauer.2008, Riihimaki.2010, Neumann.2013} or on the model increments between predefined pairs of sampling points~\cite{Chuang.2020}.

A disadvantage of the projection- and rearrangement-based approaches~\cite{Dette.2006, Dette.2006b, Lin.2014} from the point of view of manufacturing applications is that these methods are tailored to large data sets. Another disadvantage of these approaches is that the resulting models typically exhibit distinctive kinks, which are almost always unphysical. Also, the models resulting from the multidimensional rearrangement method by \cite{Dette.2006b} are not guaranteed to be monotonic when trained on small data sets. A drawback of the tilting approach from~\cite{Hall.2001} is that it is formulated and validated only for one-dimensional input spaces (intervals in $\mathbb{R}$). Accordingly, naively extending the non-adaptive discretization scheme from~\cite{Hall.2001} to higher dimensions would result in long computation times. A downside of the in-training methods from~\cite{Lauer.2008, Riihimaki.2010, Neumann.2013} is that the sampling points at which the monotonicity constraints are imposed have to be chosen in advance (even though they need not coincide with the training data points). And finally, the method from~\cite{Gupta.2016} is limited to multilinear models (that is, linear combinations of monomials $x_1^{\alpha_1} \dotsb x_d^{\alpha_d}$ where each exponent $\alpha_j$ is either $0$ or $1$). Such multilinear models are potentially not general and complex enough for various real-world applications.

The method proposed in the present article addresses the aforementioned issues and shortcomings. In Sec.~\ref{sec:methods}, our methodology for monotonic regression using semi-infinite optimization is introduced. It incorporates the monotonicity knowledge already during training. Specifically, polynomial regression models are assumed for the input-output relationships to be learned. Since there is no after-training monotonization step in the method, our models are smooth and, in particular, do not exhibit kinks. Also, due to the employed adaptive discretization scheme, the method is computationally efficient also in higher dimensions. As far as we know, such an adaptive scheme has not been applied to solve monotonic regression problems before, especially not in situations with sparse data. In Sec.~\ref{sec:results}, the method is validated by means of two applications to real-world processes which are both introduced in Sec.~\ref{sec:applications}, namely laser glass bending and press hardening of sheet metal. It turns out that the adaptive semi-infinite optimization approach to monotonic regression is better suited for the considered applications with their small data sets and the resulting models are more accurate than those obtained with the comparative approaches from the literature.

\section{Methods}\label{sec:methods}

In this section (more precisely in Secs.~\ref{sec:SIAMOR-formulation}--\ref{sec:SIAMOR-algo}), our semi-infinite optimization approach to monotonic regression is introduced. In Sec.~\ref{sec:computing-mon-proj}, our rather simple method for numerically computing the monotonic projections from~\cite{Lin.2014}, which are later compared to the models obtained with our semi-infinite method, is described.

\subsection{Monotonic regression using semi-infinite optimization} \label{sec:SIAMOR-formulation}

In our approach to monotonic regression, polynomial models
\begin{align} \label{eq:poly-regr-model}
\bm{x} \mapsto \widehat{y}_{\bm{w}}(\bm{x}) 
= \sum_{|\alpha| \le m} w_{\alpha} \bm{x}^{\alpha}
\in \mathbb{R}
\end{align}
are used for all input-output relationships $\bm{x} \mapsto y(\bm{x})$ to be learned. In the above relation~\eqref{eq:poly-regr-model}, the sum extends over all $d$-dimensional multi-indices~\cite[Sec.~1.1]{Friedlander.1998} $\alpha = (\alpha_1, \dots, \alpha_d) \in \mathbb{N}_0^d$ with degree $|\alpha| := \alpha_1 + \dotsb + \alpha_d$ less than or equal to some total degree $m \in \mathbb{N}$. The terms $\bm{x}^{\alpha} := x_1^{\alpha_1} \dotsb x_d^{\alpha_d}$ are the monomials in $d$ variables of degree less than or equal to $m$ and $w_{\alpha}$ are the corresponding model parameters to be tuned by regression. Since there are exactly 
\begin{align}
N_m=\sum_{k=0}^{m} \binom{k+d-1}{d-1} = \binom{m+d}{m}
\end{align}
$d$-dimensional monomials of degree less than or equal to $m$, the polynomial regression model~\eqref{eq:poly-regr-model} can be equivalently written as
\begin{align} \label{eq:poly-regr-model-basis-fct-enumeration}
\widehat{y}_{\bm{w}}(\bm{x}) 
= \sum_{i=1}^{N_m} w_i \phi_i(\bm{x}) = \bm{w}^{\top} \bm{\phi}(\bm{x})\,,
\end{align}
where the basis functions $\phi_1, \dots, \phi_{N_m}$ constitute any enumeration of the $d$-dimensional monomials of degree less than or equal to $m$, while $\bm{w}:=(w_1, \dots, w_{N_m})^{\top}$ and $\bm{\phi}(\bm{x}):= (\phi_1(\bm{x}), \dots, \phi_{N_m}(\bm{x}))^{\top}$.

Standard polynomial regression without regularization~\cite[Sec.~2.1.2]{Rasmussen.2006} is about solving the unconstrained optimization problem
\begin{align} \label{eq:standard-poly-regr}
\min_{\bm{w} \in \mathbb{R}^{N_m}} \frac{1}{2} \sum_{l=1}^N \big( \widehat{y}_{\bm{w}}(\bm{x}_l) - t_l \big)^2
\end{align}
or, in other words, about optimally adapting the model parameters $w_i \in \mathbb{R}$ of the polynomial model~\eqref{eq:poly-regr-model-basis-fct-enumeration} to the available data set $\mathcal{D} = \{(\bm{x}_l,t_l): l \in \{1,\dots,N\} \}$ containing $N$ points. As is well-known, the standard polynomial regression problem~\eqref{eq:standard-poly-regr}, for any given data set $\mathcal{D}$, has a unique analytical minimum-norm solution $\bm{w}$~\cite[Sec.~4.8.5]{Stoer.1993}.

In general, the resulting model $\bm{x} \mapsto \widehat{y}_{\bm{w}}(\bm{x})$ does not necessarily exhibit the monotonicity behaviour an expert expects for the underlying true physical relationship $\bm{x} \mapsto y(\bm{x})$. In order to enforce the expected monotonicity behaviour, the following constraints on the signs of the partial derivatives $\partial_{x_j} \widehat{y}_{\bm{w}}(\bm{x})$ are added to the unconstrained standard regression problem~\eqref{eq:standard-poly-regr}:
\begin{align} \label{eq:mon-constraints}
\sigma_j \cdot \partial_{x_j} \widehat{y}_{\bm{w}}(\bm{x}) \ge 0 
\quad \text{for all } j \in J \text{ and } \bm{x} \in X\,.
\end{align}
The numbers $\sigma_j \in \{-1,0,1\}$ indicate the expected monotonicity behaviour for each coordinate direction $j \in \{1,\dots, d\}$: 
\begin{itemize}
\item $\sigma_j = 1$ or $\sigma_j = -1$ indicate that $\bm{x} \mapsto y(\bm{x})$ is expected to be, respectively, monotonically increasing or decreasing in the $j$th coordinate direction,
\item $\sigma_j = 0$ indicates that one has no monotonicity knowledge in the $j$th coordinate direction. 
\end{itemize}
Also, $J := \{j \in \{1,\dots,d\}: \sigma_j \ne 0\}$ is the set of all directions for which a monotonicity constraint is imposed, and the vector 
\begin{align*}
\bm\sigma := (\sigma_1, \dots, \sigma_d)
\end{align*}
is referred to as the monotonicity signature of the relationship $\bm{x} \mapsto y(\bm{x})$. 
And finally, $X \subset \mathbb{R}^d$ is the (continuous) subset of the input space on which the polynomial model~\eqref{eq:poly-regr-model} is supposed to be a reasonable prediction for $\bm{x} \mapsto y(\bm{x})$. In this work, $X$ is chosen to be identical with the range covered by the input training data points $\bm{x}_1, \dots, \bm{x}_N$. I.e., $X$ is the compact hypercuboid set
\begin{align}
X = [a_1,b_1] \times \dotsb \times [a_d,b_d]
\end{align}
with $a_j := \min_{l=1,\dots,N} x_{l,j}$ and $b_j := \max_{l=1,\dots,N} x_{l,j}$ and with $x_{l,j}$ denoting the $j$th component of the $l$th input data point $\bm{x}_l$.
%
Writing
\begin{gather}
f(\bm{w}) := \frac{1}{2} \sum_{l=1}^N ( \widehat{y}_{\bm{w}}(\bm{x}_l) - t_l )^2 \,, \label{eq:def-obj-fct-f}\\
g_j(\bm{w},\bm{x}) := \sigma_j \cdot \partial_{x_j} \widehat{y}_{\bm{w}}(\bm{x}) \label{eq:def-constr-fcts-g_j}
\end{gather}
for brevity, our monotonic regression problem~\eqref{eq:standard-poly-regr}--\eqref{eq:mon-constraints} takes the neat and simple form
\begin{equation} \label{eq:SIP}
\begin{split}
&\min_{\bm{w}\in \mathbb{R}^{N_m}} f(\bm{w}) \\
&\quad \text{s.t.} \quad g_j(\bm{w},\bm{x}) \ge 0 \quad \text{for all } j \in J \text{ and } \bm{x} \in X\,.
\end{split}
\end{equation}
Since the input set $X$ is continuous and hence contains infinitely many points $\bm{x}$, the monotonic regression problem~\eqref{eq:SIP} features infinitely many inequality constraints. Consequently,~\eqref{eq:SIP} is a semi-infinite optimization problem~\cite{Hettich.1982, Polak.1997, Reemtsen.1998, Stein.2003, Stein.2012, Schwientek.2020} (or more precisely, a standard semi-infinite optimization problem, as opposed to a generalized one). It is well-known 
that the monotonic regression problem~\eqref{eq:SIP}, just like any other semi-infinite problem, can be equivalently rewritten as a bi-level optimization problem~\cite{Stein.2003, Shimizu.1997, Dempe.2015}, namely
\begin{equation} \label{eq:SIP-bi-level}
\begin{split}
&\min_{\bm{w}\in \mathbb{R}^{N_m}} f(\bm{w}) \\
&\quad \text{s.t.} \quad \min_{\bm{x} \in X} g_j(\bm{w},\bm{x}) \ge 0 \quad \text{for all } j \in J\,.
\end{split}
\end{equation}
Commonly, the overall optimization problem~\eqref{eq:SIP-bi-level} is referred to as the upper-level problem of~\eqref{eq:SIP}, while the subproblems in the constraints of~\eqref{eq:SIP-bi-level} are called the lower-level problems of~\eqref{eq:SIP}. It is also well-known~\cite{Stein.2003, Stein.2012} that the innocent-looking reformulation of semi-infinite problems like~\eqref{eq:SIP} as bi-level problems is the key for both the theoretical and the numerical treatment of such problems.

\subsection{Adaptive solution strategy} \label{sec:SIAMOR-strategy}

In order to solve the semi-infinite monotonic regression problem~\eqref{eq:SIP}, a variant of the adaptive, iterative discretization algorithm by \cite{Blankenship.1976} is used. In a nutshell, the idea is the following: the infinite index set $X$ of the original regression problem~\eqref{eq:SIP} is iteratively replaced by discretizations, that is, finite subsets $X^k \subset X$. These discretizations are adaptively refined from iteration to iteration. In that manner, in every iteration $k$ the ordinary (finite) optimization problem
\begin{equation} \label{eq:Appr(X^k)}
\begin{split}
&\min_{\bm{w}\in \mathbb{R}^{N_m}} f(\bm{w}) \\
&\quad \text{s.t.} \quad g_j(\bm{w},\bm{x}) \ge 0 \quad \text{for all } j \in J \text{ and } \bm{x} \in X^k
\end{split}
\end{equation}
featuring only finitely many inequality constraints is obtained. 
\eqref{eq:Appr(X^k)} is referred to as the $k$th (discretized) upper-level problem (or, following \cite{Blankenship.1976}, the $k$th approximating problem for~\eqref{eq:SIP}). Then each iteration $k$ consists of two steps, namely an optimization step and an adaptive refinement step. The optimization step computes a solution $\bm{w}^k$ of the $k$th upper-level problem~\eqref{eq:Appr(X^k)}. And the refinement step computes for each direction $j \in J$ a point $\bm{x}^{k+1,j} \in X$ at which the $j$th monotonicity constraint at $\bm{w} = \bm{w}^k$ is violated most. I.e., for every $j \in J$ an approximate solution $\bm{x}^{k+1,j}$ of the global optimization problem
\begin{align} \label{eq:Aux(w^k)}
\min_{\bm{x} \in X} g_j(\bm{w}^k,\bm{x})
\end{align}
is computed, which is referred to as the $(k,j)$th lower-level problem (or, following~\cite{Blankenship.1976}, the $(k,j)$th auxiliary problem). Then all the points $\bm{x}^{k+1,j}$ for which a monotonicity violation occurs are added to the current discretization $X^k$ in order to obtain the new discretization $X^{k+1}$. As soon as no more monotonicity violations occur, iterating is stopped. 

With regard to the practical implementation of the above solution strategy, it is important to observe that the discretized upper-level problems~\eqref{eq:Appr(X^k)} are standard convex quadratic programs~\cite{Nocedal.2006}. Indeed, inserting~\eqref{eq:poly-regr-model-basis-fct-enumeration} into~\eqref{eq:def-obj-fct-f} and using the design matrix $\Phi$ with entries $\Phi_{li} := \phi_i(\bm{x}_l)$, one obtains
\begin{align*}
f(\bm{w}) = \frac{1}{2} \|\Phi \bm{w} - \bm{t}\|_2^2 =
\frac{1}{2} \bm{w}^{\top} \Phi^{\top}\Phi \bm{w} - \bm{t}^{\top} \Phi \bm{w} + \frac{1}{2} \bm{t}^{\top} \bm{t}\,.
\end{align*}
Consequently, the objective function of~\eqref{eq:Appr(X^k)} is indeed quadratic and convex w.r.t.~$\bm{w}$. Additionally, in view of
\begin{align} \label{eq:constr-linear-in-w}
g_j(\bm{w},\bm{x}) = \sigma_j \cdot \partial_{x_j} \widehat{y}_{\bm{w}}(\bm{x}) = \sigma_j \cdot \bm{w}^{\top} \big( \partial_{x_j} \bm{\phi}(\bm{x}) \big)\,,
\end{align}
the constraints of~\eqref{eq:Appr(X^k)} are indeed linear w.r.t.~$\bm{w}$. 

With regard to the practical implementation, it is also important to observe that the objective functions $\bm{x} \mapsto g_j(\bm{w}^k,\bm{x})$ of the lower-level problems~\eqref{eq:Aux(w^k)} are non-convex polynomials and therefore in general have several local minima. So, \eqref{eq:Aux(w^k)} needs to be solved numerically with a global optimization solver.

\subsection{Algorithm and implementation details}\label{sec:SIAMOR-algo}

In the following, our adaptive discretization algorithm is described in detail. As has already been pointed out above, it is a variant of the general algorithm developed by \cite[Sec.~2]{Blankenship.1976}, and it is explained after Algorithm~1 how our variant differs from its prototype~\cite{Blankenship.1976}.
%
\begin{algo}
\begin{enumerate}
\item Initialize: set $k = 0$ and choose $X^0$ as a coarse (but non-empty) rectangular grid in $X$.
\item Solve the $k$th upper-level problem~\eqref{eq:Appr(X^k)} to obtain optimal model parameters $\bm{w}^k \in \mathbb{R}^{N_m}$.
\item Solve the $(k,j)$th lower-level problem~\eqref{eq:Aux(w^k)} approximately for every $j \in J$ to find approximate global minimizers $\bm{x}^{k+1,j} \in X$. Add those of the points $\bm{x}^{k+1,j}$, for which substantial monotonicity violations occur, i.e., for which $g_j(\bm{w}, \bm{x}^{k+1,j}) < -\varepsilon_j$, to the current discretization $X^k$ and go to Step 2 with $k = k+1$. If for none of the points $\bm{x}^{k+1,j}$ substantial monotonicity violations occur, go to Step~4. 
\item Check for monotonicity violations on a fixed, fine rectangular reference grid $X_{\mathrm{ref}} \subset X$. If there are no such violations, that is, if $g_j(\bm{w}^k,\bm{x}) \ge-\varepsilon_j$ for all $j \in J$ and $\bm{x} \in X_{\mathrm{ref}}$, stop. Else, for every direction $j$ with violations, add the reference grid point $\bm{x}^{k+1,j}_{\mathrm{ref}}$ with the largest violation to $X^k$ and go to Step 2 with $k = k+1$. 
\end{enumerate}
\end{algo}

In contrast to \cite{Blankenship.1976}, the Algorithm~1 above does not require exact solutions of the (non-convex) lower-level problems. Indeed,  Step~3 of Algorithm~1 only requires to approximate a solution numerically. Therefore, slight constraint violations are tolerated and the finalization step (Step~4) is introduced. Another, but minor,  difference compared to the algorithm from~\cite{Blankenship.1976} is that there there are several lower-level problems in each iteration and not just one, because monotonicity is enforced in multiple coordinate directions in general.

As for the tolerances $\varepsilon_j$ (Steps~3 and 4 of Algorithm~1), a monotonicity violation of 1\,\% of the ranges covered by the in- and output training data is allowed for:
\begin{align} \label{eq:eps_j-def}
\varepsilon_j = 0.01 \frac{\max_{l=1,\dots,N}t_l - \min_{l=1,\dots,N}t_l}{\max_{l=1,\dots,N} x_{l,j} - \min_{l=1,\dots,N} x_{l,j}}\,.
\end{align}
The degrees $m$ of the polynomial models in this work are chosen as the largest possible values that do not result in an overfit, because increasing $m$ enhances the model accuracy in general. In this respect, the number of model parameters is allowed to exceed the number of data points ($N_m\geq N$), since the constraints represent additional information supplementing the data. As for the reference grid $X_\mathrm{ref}$ in the finalization step (Step~4 of Algorithm~1), 20 values per input dimension equidistantly distributed from the lower to the upper bound along each direction are used. 

Algorithm~1 was implemented in Python and the package sklearn was used for the numerical representation of the models. Since the discretized upper-level problems~\eqref{eq:Appr(X^k)} are standard convex quadratic programs, a solver tailored to that specific problem class is used, namely quadprog~\cite{Goldfarb.1983}. It can solve quadratic programs with hundreds of variables and thousands of constraints in just a few seconds because it efficiently exploits the simple structure of the problem. Since on the other hand the lower-level problems~\eqref{eq:Aux(w^k)} are global optimization problems with possibly several local minima, a suitable global optimization solver is required. The solver scipy.optimize.shgo~\cite{Endres.2018} was used, which employs a simplicial homology strategy and which, in our applications, turned out to be a good compromise between speed and reliability. For the problems considered in this article, shgo's internal local optimization was configured to occur in every iteration, to multi-start from a Sobol set of 100$\times d$ points and to be executed using the algorithm L-BFGS-B with analytical gradients. Step~4 in Algorithm~1 ensures that shgo does not miss locations where the monotonicity is not as required.

\subsection{Computing monotonic projections} \label{sec:computing-mon-proj}

In order to validate our semi-infinite optimization approach to monotonic regression, it will be compared, among other things, to the projection-based monotonization approach by \cite{Lin.2014}. As has already been pointed out in Sec.~\ref{sec:intro}, the projection method starts out from a purely data-based initial model $\widehat{y}^{\,0}$ (a Gaussian process regressor in the case of~\cite{Lin.2014}) and then replaces this initial model by the monotonic projection $\widehat{y}$ of $\widehat{y}^{\,0}$. I.e., $\widehat{y}: X \to \mathbb{R}$ is the monotonic square-integrable function with monotonicity signature $\bm\sigma$ that is closest to $\widehat{y}^{\,0}$ in the $L^2$-norm. 

In order to numerically compute this monotonic projection $\widehat{y}$, the original procedure proposed by \cite{Lin.2014} is not used here, though. Instead, the conceptually simpler methodology by \cite{Schmid.2020} is employed. First, the input space $X$ is discretized with a fine rectangular grid $G$. Then the corresponding discrete monotonic projection $(\widehat{y}(\bm{x}))_{\bm{x}\in G}$, that is, the solution of the constrained optimization problem
\begin{equation} \label{eq:discr-mon-proj}
\begin{split}
&\min_{z \in \mathbb{R}^G} \sum_{\bm{x}\in G} \big( z(\bm{x})-\widehat{y}^{\,0}(\bm{x}) \big)^2 \\
&\quad \text{s.t.} \quad \sigma_j \cdot \big( z(\bm{x}+h_j \bm{e}_j) - z(\bm{x}) \big) \ge 0 \quad \text{for all } j \in J \\
&\quad \text{and all } \bm{x} \in G \text{ for which } \bm{x}+h_j\bm{e}_j\in G \,,
\end{split}
\end{equation}
is computed. In the above relation, $\mathbb{R}^G$ is the $|G|$-dimensional vector space of all $\mathbb{R}$-valued functions $z=(z(\bm{x}))_{\bm{x}\in G}$ defined on the discrete set $G$, $h_j>0$ indicates the distance of adjacent grid points in the $j$th coordinate direction, and $\bm{e}_j \in \mathbb{R}^d$ is the $j$th canonical unit vector. It is shown in~\cite{Schmid.2020} that the extension of $(\widehat{y}(\bm{x}))_{\bm{x}\in G}$ to a grid-constant function on the whole of $X$ is a good approximation of the monotonic projection $\widehat{y}$, if only the grid is fine enough and the initial model $\widehat{y}^{\,0}$ is continuous, for instance.

Since both the objective function and the constraints of~\eqref{eq:discr-mon-proj} are convex w.r.t.~$z$, the problem~\eqref{eq:discr-mon-proj} is a convex program. cvxopt~\cite{Andersen.2011} is used to solve these problems because it offers a sparse matrix type to represent the large coefficient and constraint matrices for $d>1$. Alternatively, the discrete monotonic projection problems can also be solved using any of the more sophisticated computational methods by \cite[Sec.~2.3]{Barlow.1972}; \cite[Sec.~4.1]{Robertson.1988}; \cite{Qian.1996, Spouge.2003, Stout.2013, Stout.2015} or \cite{Kyng.2015}. Yet, for the number of input dimensions considered here, our direct computational method is sufficient.

\section{Applications in manufacturing}\label{sec:applications}

\subsection{Laser glass bending} \label{sec:glass-bending}

\begin{figure}[htbp]
	\centering
	\includegraphics[width=0.45\columnwidth]{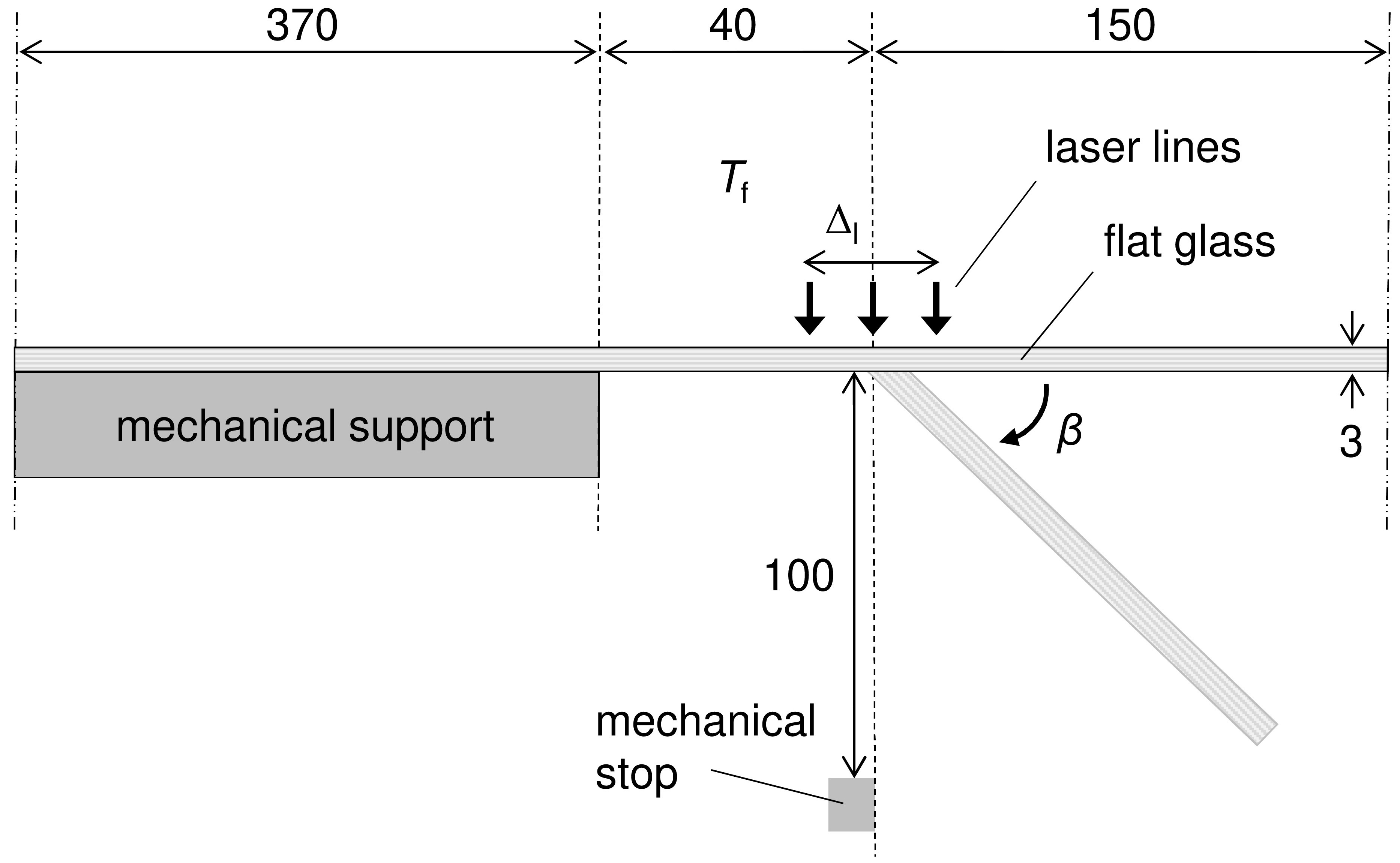}
	\caption{Side view of the laser glass bending process. Symbols: $T_{\mathrm{f}}$ -- furnace temperature, $\Delta_{\mathrm{l}}$ -- distance between the left- and right-most laser line, $\beta$ -- bending angle. Lengths are given in mm.}
	\label{fig:glass-bending_illustration}
\end{figure}

The first application example is laser glass bending. In the industrial standard process of glass bending~\cite{Neugebauer.2014}, a flat glass specimen is placed in a furnace with furnace temperature $T_{\mathrm{f}}$, and then the heated specimen is bent at a designated edge driven by gravity. As an additional feature, a laser can be added to the industrial standard process in order to specifically heat up the critical region of the flat glass around the bending edge and thus, to speed up the process and achieve smaller bending radii~\cite{Rist.2018, Rist.2019}. The laser can generally scribe in arbitrary directions. In the process considered here, however, the laser path is restricted to three straight lines parallel to the bending edge. While the middle line defines the bending edge, the outer two lines are at a fixed distance $\Delta_{\mathrm{l}}/2 =5.75\,$ mm in each direction to it. The laser spot moves along this path in multiple cycles with the number of cycles denoted by $n_{\mathrm{c}}$. The scribing speed and the power of the laser are held constant. A mechanical stop below the bending edge guarantees that the bending angle does not exceed $90^\circ$. An illustration of the laser glass bending process is shown in Fig.~\ref{fig:glass-bending_illustration}.

The goal of the glass bending process considered here is to obtain bent glass parts with a bending angle as close as possible to $90^\circ$. In order to achieve this goal, a sufficient amount of heat has to be induced. Thus, the modelled quantity is the bending angle $y := \beta$ as a function of the two process variables
\begin{align} \label{eq:process-var-glass-bending}
\bm{x} := (x_1,x_2) := (T_{\mathrm{f}}, n_{\mathrm{c}}) \in X\,,
\end{align} 
where $X \subset \mathbb{R}^2$ is the rectangular set with the bounds specified in Tab.~\ref{tab:glass-bending_paramrange}.

\begin{table}[htbp]
	\centering
	\caption{Ranges for the process variables of laser glass bending}
	\label{tab:glass-bending_paramrange}
	\begin{tabular*}{0.6\columnwidth}{cccc}
		\hline\hline
		Variable\hspace{1em} & \hspace{1em}Min\hspace{1em} & \hspace{1em}Max\hspace{1em} & \hspace{1em}Phys. unit \\
		\hline
		$T_{\mathrm{f}}$ & $480$ & $560$ & $^\circ$C \\
		$n_{\mathrm{c}}$ & $40$ & $50$ & -- \\
		\hline\hline
	\end{tabular*}
\end{table}

Since generating experimental training data from the real process is cumbersome, a two-dimensional finite-element model was set up to generate data numerically. The simulation of the process is based on a coupled thermo-mechanical problem with finite deformation. Since the CO$_2$ laser used in the process operates in the opaque wave length spectrum of glass, the heat supply is modelled as a surface flux into the deforming sheet. In this two-dimensional setting, the heat is assumed to be deposited instantaneously along the thickness direction and also instantaneously on all three laser lines. Radiation effects are ignored here and heat conduction inside the glass is described by the classical Fourier law with the heat conductivity obtained experimentally via laser flash analysis. In view of the relevant relaxation and process time scales for the applied temperature range, the mechanical behaviour of the glass is described by a simple Maxwell-type visco-elastic law.
The deformation due to gravity is heavily affected by the pronounced temperature dependence of the viscosity above the glass transition, which is described here using the Williams-Landel-Ferry approximation~\cite{Williams.1955}.
The simulation is conducted using the commercial finite-element code Abaqus\textsuperscript{\textcopyright}. It was used to create a training data set comprising 25 data points sampled on a 2D rectangular grid. The  values used for the two degrees of freedom (five for $T_{\mathrm{f}}$ and five for $n_{\mathrm{c}}$) are placed equidistantly from the lower to the upper bounds given in Tab.~\ref{tab:glass-bending_paramrange}. 

Within these ranges and for the laser configuration described above, process experts expect the following monotonicity behaviour: the bending angle $y = \beta$ increases monotonically with increasing glass temperature in the critical region and thus, with increasing $T_{\mathrm{f}}$ and $n_{\mathrm{c}}$. In other words, the monotonicity signature $\bm\sigma$ of the bending angle $y$ as a function of the inputs $\bm{x}$ from~\eqref{eq:process-var-glass-bending} is expected to be
\begin{align} \label{eq:mon-signature-glass-bending}
\bm\sigma = (\sigma_1,\sigma_2) = (1,1).
\end{align}

\subsection{Forming and press hardening of sheet metal} \label{sec:press-hardening}

\begin{figure}[htbp]
	\centering
	\includegraphics[width=0.45\columnwidth]{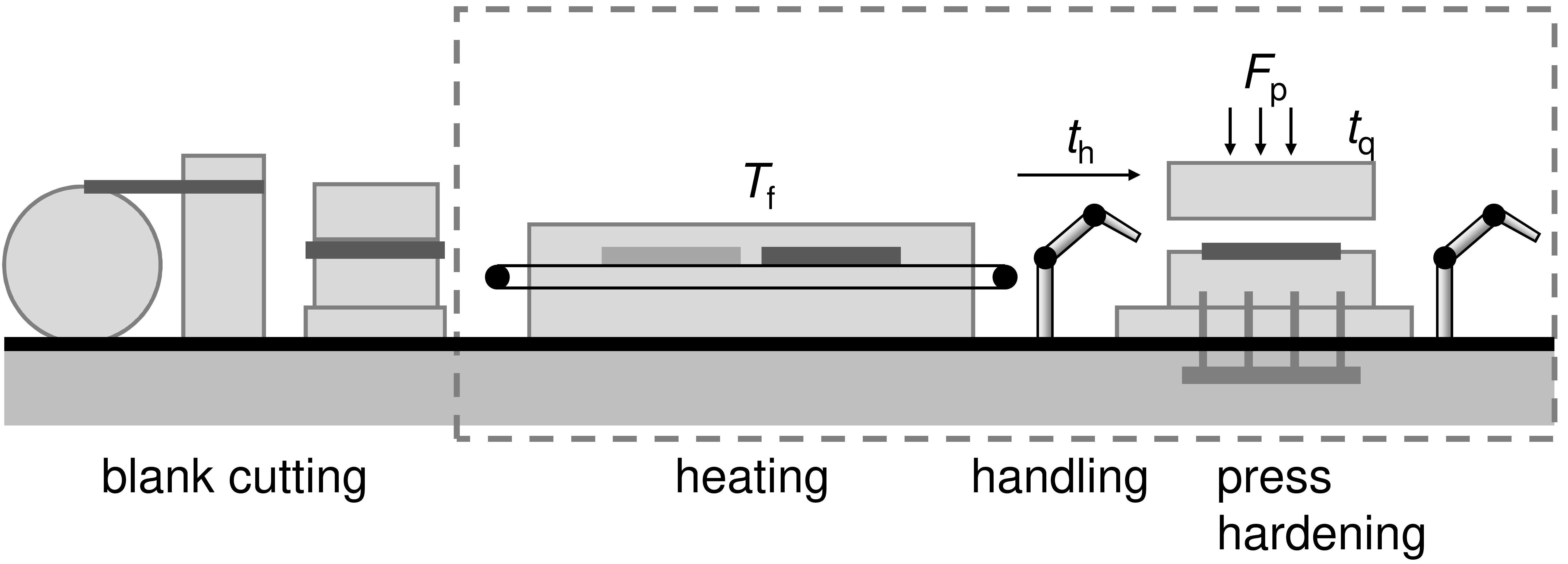}
	\caption{Side view of the press hardening process~\cite{Neugebauer.2012} indicating the considered process steps. Symbols: $T_{\mathrm{f}}$ -- furnace temperature, $t_{\mathrm{h}}$ -- handling time, $F_{\mathrm{p}}$ -- press force, $t_{\mathrm{q}}$ -- quenching time.}
	\label{fig:press-hardening}
\end{figure}

Another application example is press hardening~\cite{Neugebauer.2012}. Within the used experimental setup of this process, a blank is placed in a chamber furnace with a furnace temperature $T_{\mathrm{f}}$ above $900\,^\circ$C. After heating the blank, an industrial robot transports it with handling time $t_{\mathrm{h}}$ into the cooled forming tool. In the following, the extra handling time \mbox{$\Delta t_{\mathrm{h}}=t_{\mathrm{h}}-10\,$s} is used instead, with $10\,$s being the minimum time the robot takes to move the blank from the furnace to the press. The final combined forming and quenching step allows for the variation of the press force $F_{\mathrm{p}}$ and the quenching time $t_{\mathrm{q}}$. Afterwards, the formed part is transferred by the industrial robot to a deposition table for further cooling. An illustration of the process chain is shown in Fig.~\ref{fig:press-hardening}.

The goal of the press hardening process considered in this work is to obtain a formed metal part that is as hard as possible, where the hardness is measured in units of the Vickers hardness number (unit symbol HV). In order to achieve this goal, a sufficiently fast cooling rate during the quenching step is necessary to induce a microstructural phase change in the material, which in turn leads to high hardness values.  Thus, the output quantity to be modelled is the hardness $y$ of the formed part (at distinguished measurement points on the surface of the part) as a function of the four process variables
\begin{align} \label{eq:process-var-press-hardening}
\bm{x} := (x_1,\dots,x_4) := (T_{\mathrm{f}},\Delta t_{\mathrm{h}}, F_{\mathrm{p}}, t_{\mathrm{q}}) \in X\,,
\end{align}
where $X \subset \mathbb{R}^4$ is the hypercuboid set with the bounds specified in Tab.~\ref{tab:press-hardening_paramrange}.

\begin{table}[htbp]
	\centering
	\caption{Ranges for the process variables of press hardening}
	\label{tab:press-hardening_paramrange}
	\begin{tabular*}{0.6\columnwidth}{cccc}
		\hline\hline
		Variable \hspace{1em} & \hspace{1em}Min\hspace{1em} & \hspace{1em}Max\hspace{1em} & \hspace{1em}Phys. unit \\
		\hline
		$T_{\mathrm{f}}$ & 871 & 933 & $^\circ$C \\
		$\Delta t_{\mathrm{h}}$ & 0 & 4 & s \\
		$F_{\mathrm{p}}$ & 1750 & 2250 & kN \\
		$t_{\mathrm{q}}$ & 2 & 6 & s \\
		\hline\hline
	\end{tabular*}
\end{table}

As in the case of glass bending, however, experiments for the press hardening process are expensive because they usually require manual adjustments, which tend to be time-consuming. And the local hardness measurements at the chosen measurement points on the surface of the quenched part are time-consuming as well. This is why the training data base is rather small. It contains 60 points resulting from a design of experiments with the four process variables $T_{\mathrm{f}}$, $\Delta t_{\mathrm{h}}$, $F_{\mathrm{p}}$, $t_{\mathrm{q}}$ ranging between the bounds in Tab.~\ref{tab:press-hardening_paramrange}, along with the corresponding hardness values at six local measurement points (referred to as MP1, $\dots$, MP6 in the following).

In order to compensate this data shortage, expert knowledge is brought into play. An expert for press hardening expects the hardness to decrease monotonically with $\Delta t_{\mathrm{h}}$ and to increase monotonically with $T_{\mathrm{f}}$ as well as with $t_{\mathrm{q}}$. 
In other words, the monotonicity signature $\bm\sigma$ of the hardness $y$ (at any given measurement point) as a function of the inputs $\bm{x}$ from~\eqref{eq:process-var-press-hardening} is expected to be
\begin{align} \label{eq:mon-signature-press-hardening}
\bm\sigma = (\sigma_1,\dots,\sigma_4) = (1,-1,0,1).
\end{align}
In fact, a press hardening expert expects even a bit more, namely that the hardness grows in a sigmoid-like manner with $T_{\mathrm{f}}$ and that it grows concavely towards saturation with increasing $t_{\mathrm{q}}$. All these requirements result from qualitative physical considerations and are supported by empirical experience.

\section{Results and discussion} \label{sec:results}

In this section, the semi-infinite monotonic regression method is applied to the industrial processes described in Sec.~\ref{sec:applications} and compared to other approaches for incorporating monotonicity knowledge, which are known from the literature. The acronym SIAMOR is used for the approach. It stands for semi-infinite optimization using an adaptive discretization scheme for monotonic regression.

\subsection{Informed machine learning models for laser glass bending}

\begin{figure}[htbp]
\centering
\begin{subfigure}[b]{0.45\textwidth}
\centering
\includegraphics[width=\columnwidth]{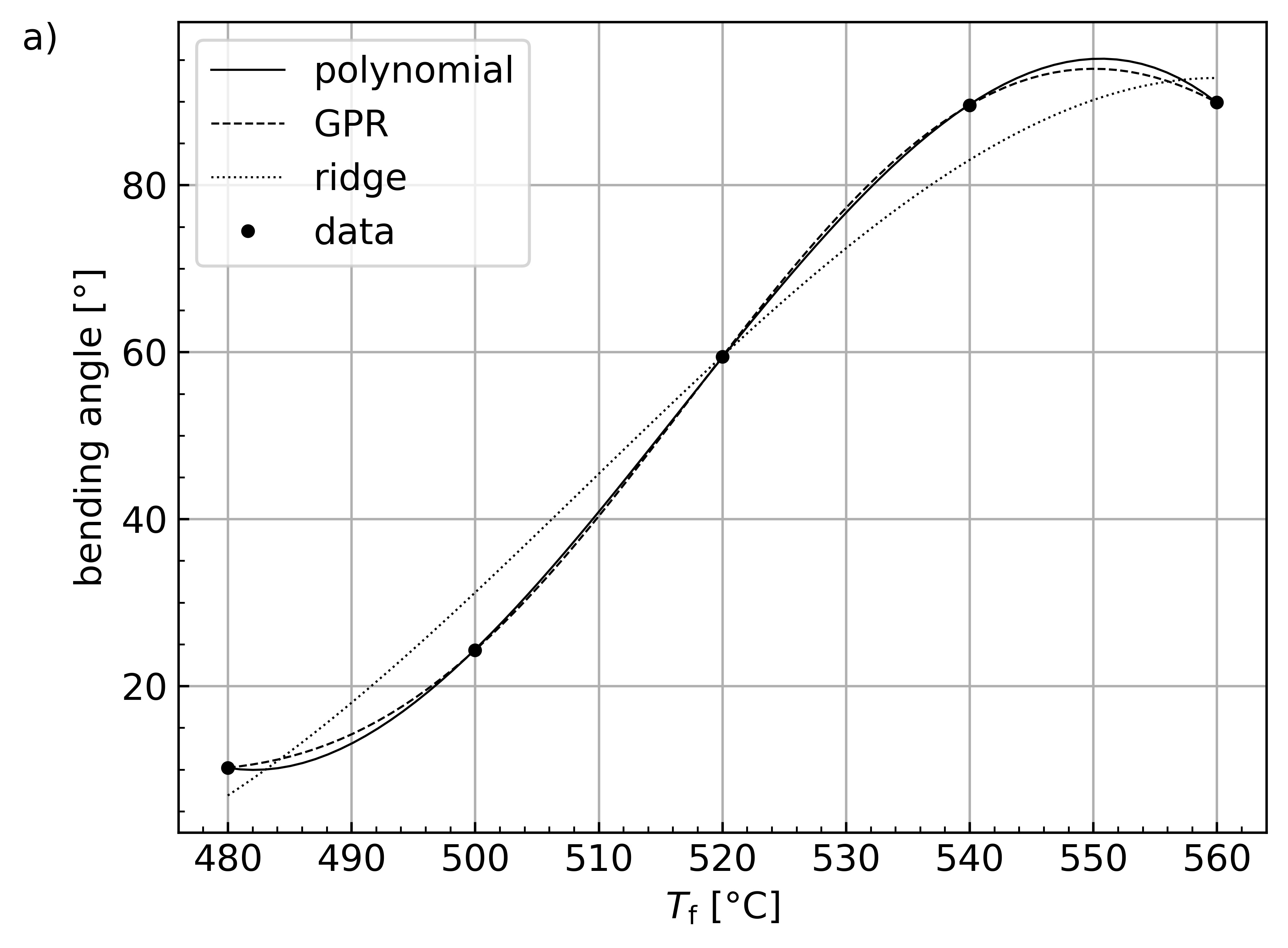}
\end{subfigure}
\begin{subfigure}[b]{0.45\textwidth}
\includegraphics[width=\columnwidth]{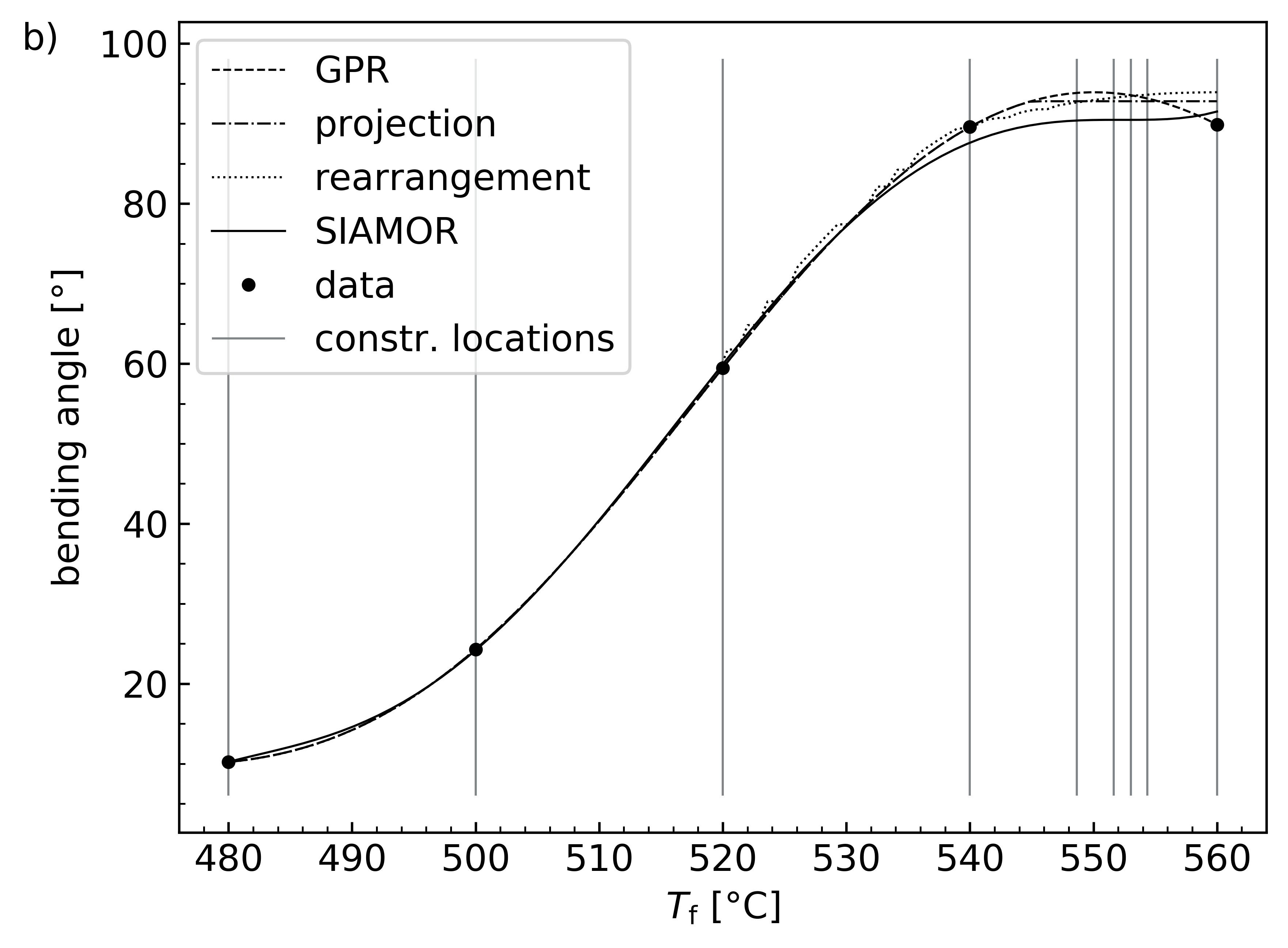}
\end{subfigure}
\caption{1D regression for laser glass bending ($n_{\mathrm{c}}=50$). a)~Unconstrained regression, solid: polynomial model ($m=3$), dashed: Gaussian process regression (GPR) with RBF kernel (noise level 10$^{-5}$), dotted: polynomial ridge regression ($m=3$, $\lambda=0.003$). b)~Monotonic regression, with the solid line resulting from the SIAMOR method (see Secs.~\ref{sec:SIAMOR-formulation}--\ref{sec:SIAMOR-algo}) with degree $m=5$. The projection~\cite{Lin.2014} (dash-dotted) and rearrangement~\cite{Dette.2006} (dotted) methods were fed with the dashed GPR curve as non-monotonic reference predictor.}
\label{fig:IWM1D}
\end{figure}

To begin with, the SIAMOR method is validated on a 1D subset of the data for laser glass bending, namely the subset of all data points for which $n_{\mathrm{c}}=50$. This means that, out of the 25 data points, five points remain for training. 
First of all, ordinary unconstrained regression techniques are tried (see Fig.~\ref{fig:IWM1D}a). A polynomial model of degree $m=3$ (solid line) and a Gaussian process regressor~\cite{Rasmussen.2006} (GPR, dashed line) do not comply with the monotonicity knowledge at high $T_{\mathrm{f}}$. A radial basis function (RBF) kernel was used for the GPR. This non-parametric model is always a reasonable choice for simulated data because it accurately reproduces the data themselves if the noise-level parameter is kept small. For all GPR models in this work, that parameter was set to 10$^{-5}$. Next, the polynomial model is regularized in a ridge regression (dotted line), where the squared $l^2$-norm $\lambda \|\bm{w}\|_2^2$ with a regularization weight $\lambda$ is added to the objective function in~\eqref{eq:standard-poly-regr}. $\lambda=0.003$ is chosen, which is roughly the minimum necessary value to achieve monotonicity. But the resulting model does not predict the data very well. Thus, all three models from Fig.~\ref{fig:IWM1D}a are unsatisfactory.

As a next step, the monotonicity requirement w.r.t.~$T_{\mathrm{f}}$ is brought to bear, and monotonic regression with the SIAMOR method ($m=5$) is used (see Fig.~\ref{fig:IWM1D}b) and compared to the rearrangement~\cite{Dette.2006} and to the monotonic projection~\cite{Lin.2014} of the Gaussian process regressor from Fig.~\ref{fig:IWM1D}a. As mentioned before, both comparative methods are based on a non-monotonic pre-trained reference predictor. This is a fundamental difference to the SIAMOR method, which imposes the monotonicity already in the training phase. The projection was calculated as described in Sec.~\ref{sec:computing-mon-proj} with $|G|=80$ grid points. For the rearrangement method, the R package monreg was invoked from Python using the package rpy2. The degree $m$ of the polynomial ansatz~\eqref{eq:poly-regr-model} used in the SIAMOR method is chosen as described in Sec.~\ref{sec:SIAMOR-algo}. For the specific case considered here, the curve starts to vary unreasonably (albeit still monotonically) between the data points for $m\geq 6$ and therefore $m=5$ was chosen. The SIAMOR algorithm was initialized with five equidistant constraint locations in $X^0$, and it converged in iteration 5 with a total number of 9 constraints. The locations of the constraints are marked in Fig.~\ref{fig:IWM1D}b by the gray, vertical lines. The adaptive algorithm automatically places the non-initial constraints in the non-monotonic region at high $T_{\mathrm{f}}$. In terms of the root-mean-squared error
\begin{align} \label{eq:RMSE}
\text{RMSE}=\sqrt{\frac 1N\sum\limits_{l=1}^N\bigl(\widehat{y}(\bm x_l) - t_l\bigr)^2}
\end{align}
on the training data, the SIAMOR model fits the data best, see Tab.~\ref{tab:IWM1D_RMSE}. Another advantage of the SIAMOR model is that it is continuously differentiable.

\begin{table}[htbp]
\centering
\caption{Root-mean-squared deviations (RMSE) of the monotonic regression models from the training data for laser glass bending (1D)}
\label{tab:IWM1D_RMSE}
\begin{tabular*}{0.6\columnwidth}{ll}
\hline\hline
Monotonic regression type & \hspace{3.5em}RMSE [$^\circ$] \\
\hline
projection \cite{Lin.2014} & \hspace{3.5em}1.3822 \\
rearrangement \cite{Dette.2006} & \hspace{3.5em}1.8432 \\
SIAMOR & \hspace{3.5em}1.1598\\
\hline\hline
\end{tabular*}
\end{table}

\begin{figure}[htpb]
\centering
\begin{subfigure}[b]{0.45\textwidth}
\centering
\includegraphics[width=\columnwidth]{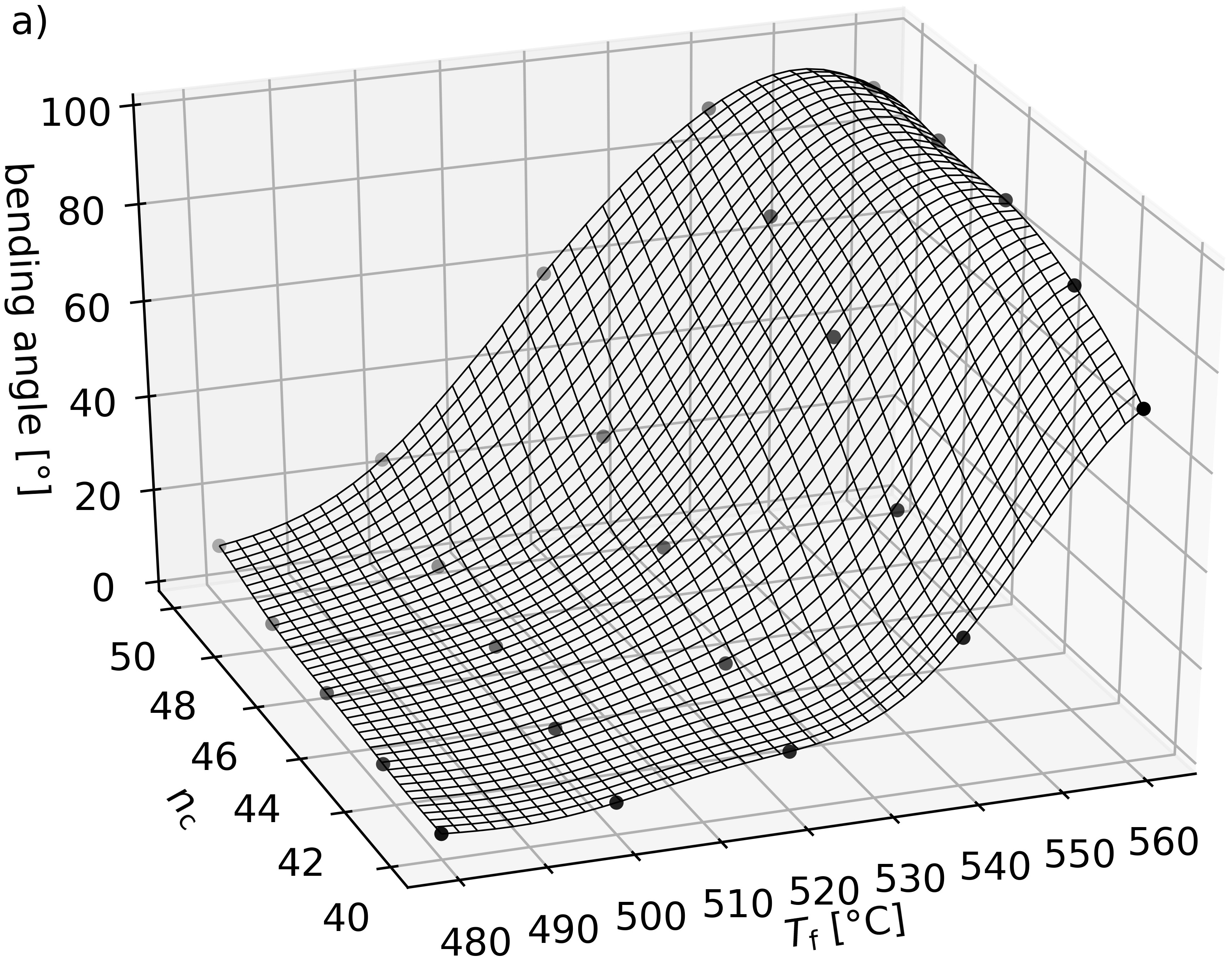}
\end{subfigure}
\hfill
\begin{subfigure}[b]{0.45\textwidth}
\centering
\includegraphics[width=\columnwidth]{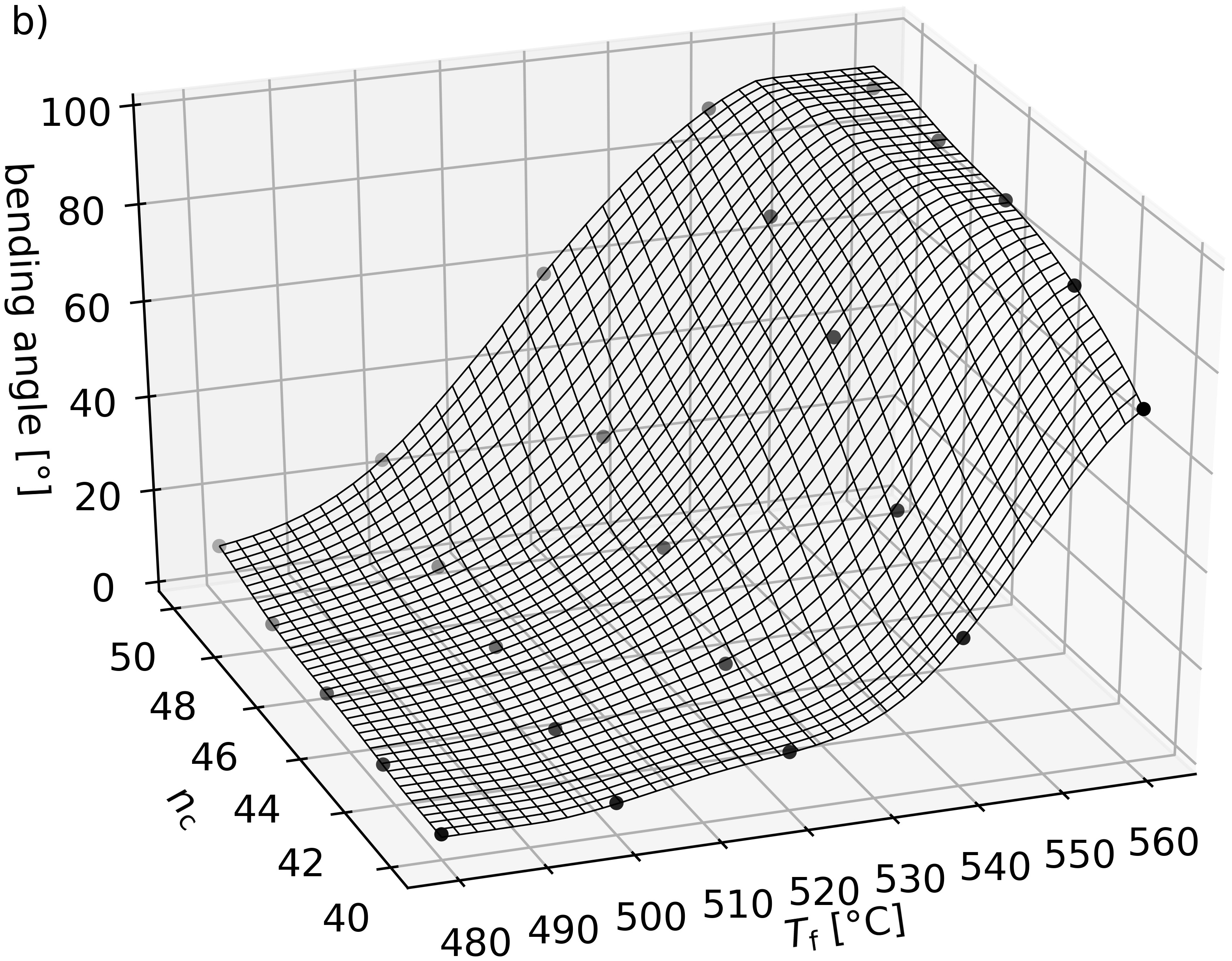}
\end{subfigure}
\vspace{0.5cm}
\begin{subfigure}[b]{0.45\textwidth}
\centering
\includegraphics[width=\columnwidth]{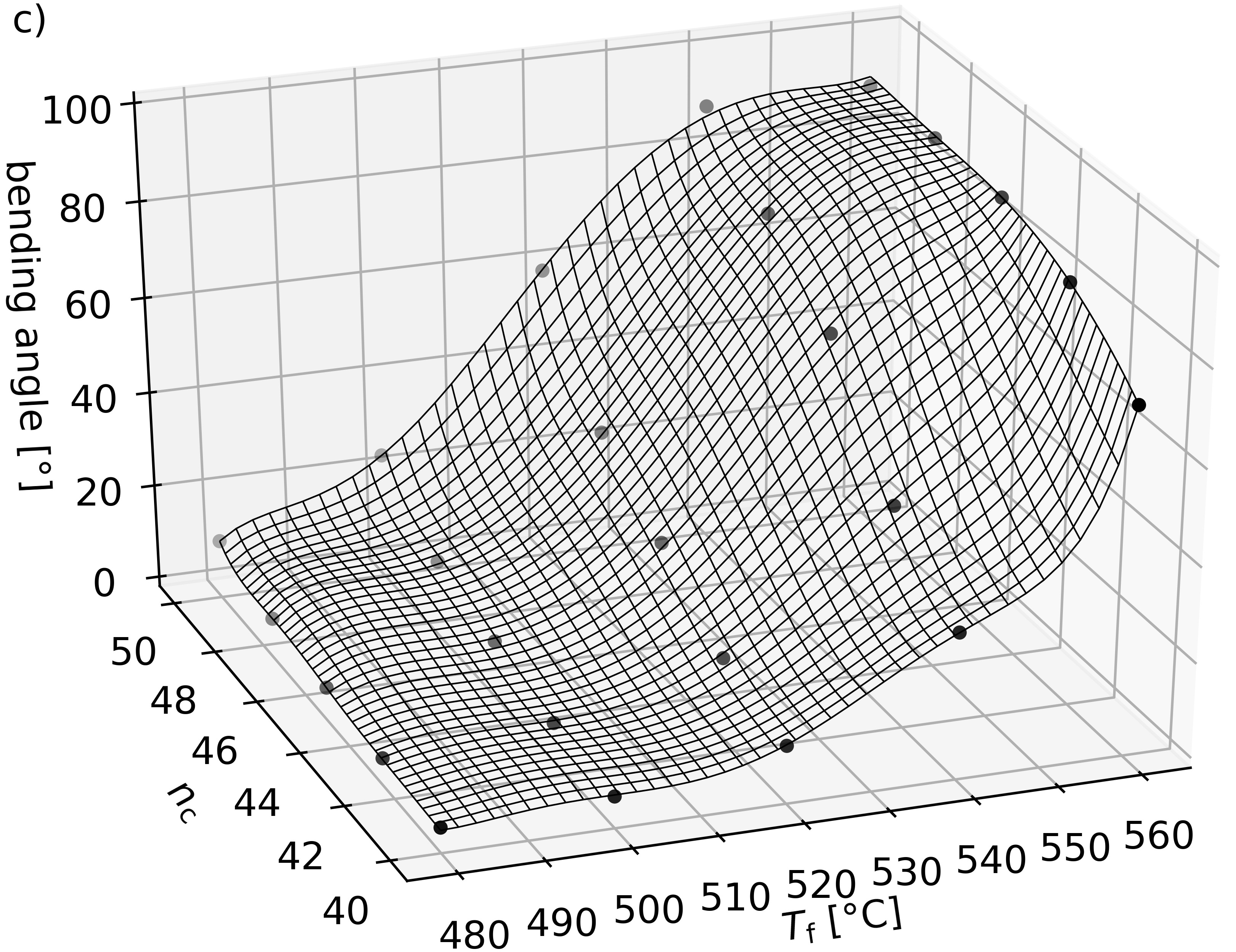}
\end{subfigure}
\caption{2D regression for laser glass bending, where the markers represent the employed training data. a)~Gaussian process regression (non-monotonic) with a multi-length-scale RBF kernel (noise level 10$^{-5}$), b)~projection~\cite{Lin.2014} of GPR, c)~monotonic regression of a polynomial model ($m=7$) using the SIAMOR method (see Secs.~\ref{sec:SIAMOR-formulation}--\ref{sec:SIAMOR-algo}).}
\label{fig:IWM2D}
\end{figure}

After these calculations on a 1D subset, the full 2D data set of the considered laser glass bending process with its 25 data points is now used. The results are shown in Fig.~\ref{fig:IWM2D}. Again, part a) of the figure displays an unconstrained Gaussian process regressor for comparison. The RBF kernel contained one length scale parameter per input dimension, and sklearn correctly adjusted these hyperparameters using L-BFGS-B. I.e., the employed length scales maximize the log-likelihood function of the model. Nevertheless, the model is unsatisfactory because it exhibits a bump in the rear right corner of the plot, contradicting the monotonicity knowledge.

Fig.~\ref{fig:IWM2D}b shows the 2D monotonic projection of the GPR with the monotonicity requirements~\eqref{eq:mon-signature-glass-bending} w.r.t.~$T_{\mathrm{f}}$ and $n_{\mathrm{c}}$. It was calculated according to Sec.~\ref{sec:computing-mon-proj} on a rectangular grid $G$ consisting of 40$^2$ points (40 values per input dimension). The resulting model looks generally reasonable and, in particular, satisfies the monotonicity specifications, but it exhibits kinks and plateaus. The most conspicuous kink starts at about $T_{\mathrm{f}}=546\,^\circ$C, $n_{\mathrm{c}}=50$ and proceeds towards the front right. The rearrangement method by \cite{Dette.2006b} is not used for comparison here because for small data sets in $d>1$, it does not guarantee monotonicity.

Fig.~\ref{fig:IWM2D}c displays the corresponding response surface of a polynomial model of the form \eqref{eq:poly-regr-model} with degree $m=7$  trained with SIAMOR. For $m=7$ there are $N_m=36$ model parameters. The discretization $X^0$ was initialized with a rectangular grid using five equidistant values per dimension. The algorithm converged in iteration 11 with 69 final constraints. The resulting model is smoother than the one in Fig.~\ref{fig:IWM2D}b and it predicts the training data more accurately. Indeed, the corresponding RMSE values are 1.2518$^{\circ}$ for projection and 0.6607$^{\circ}$ for SIAMOR.

\subsection{Informed machine learning models for forming and press hardening}\label{sec:AM3results}

\begin{figure}
\centering
\includegraphics[width=0.5\columnwidth]{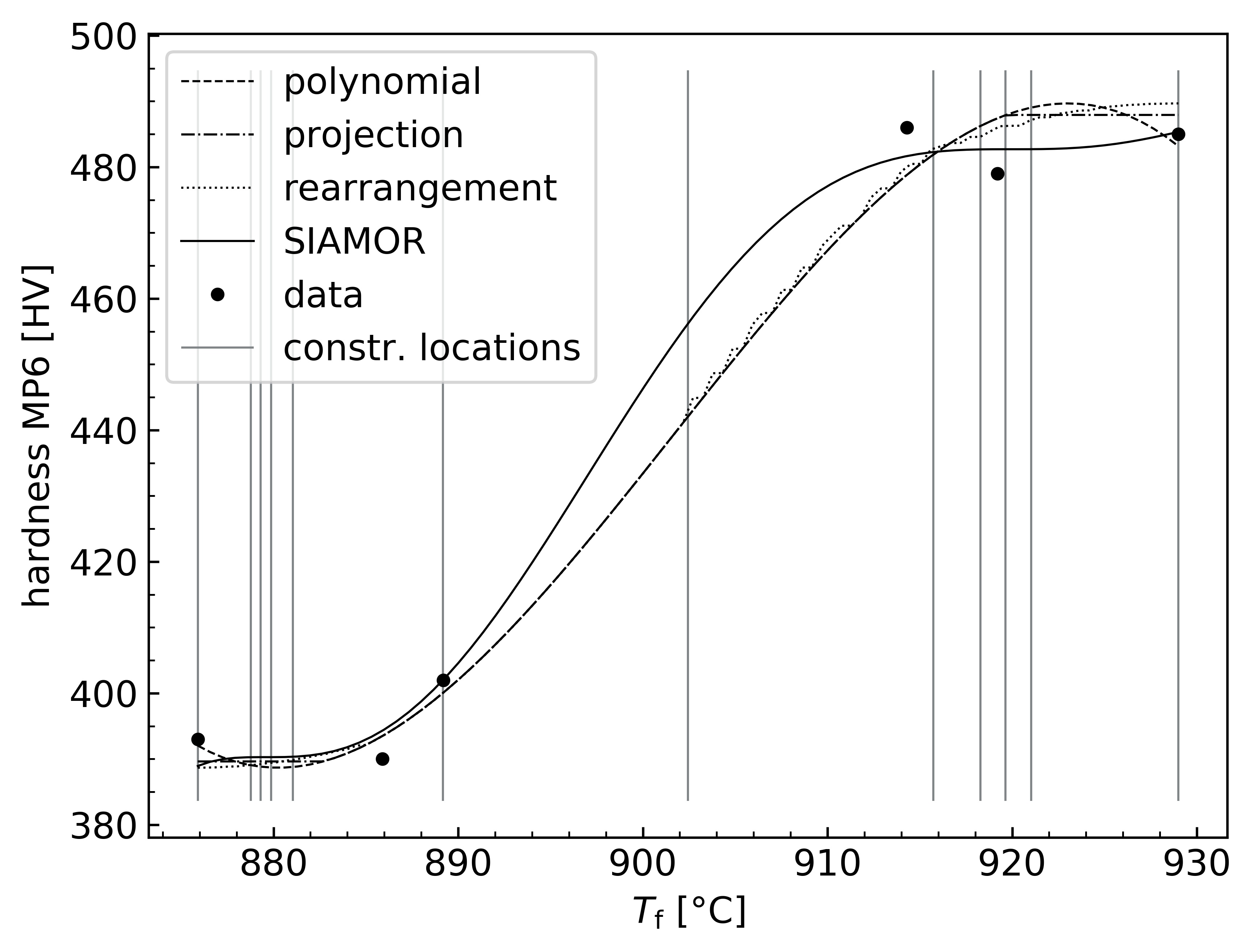}
\caption{1D regression for forming and press hardening of sheet metal ($F_{\mathrm{p}}=2250\,$kN, $\Delta t_{\mathrm{h}}=4\,$s, $t_{\mathrm{q}}=2\,$s). Dashed: (non-monotonic) polynomial of degree $m=3$ as reference model, dash-dotted: projection~\cite{Lin.2014}, dotted: rearrangement~\cite{Dette.2006}, solid: SIAMOR (see Secs.~\ref{sec:SIAMOR-formulation}--\ref{sec:SIAMOR-algo}) with degree $m=6$. The projection and rearrangement methods were fed with the dashed polynomial curve as non-monotonic reference predictor.}
\label{fig:IWU1D}
\end{figure}

As in the glass bending case, the SIAMOR method is first validated on a 1D subset of the data for the press hardening process. Namely, only those data points with $F_{\mathrm{p}}=2250\,$kN, $\Delta t_{\mathrm{h}}=4\,$s and $t_{\mathrm{q}}=2\,$s are considered. These specifications are met by six data points, and these were used to train the models shown in Fig.~\ref{fig:IWU1D}. The data reflect the expected sigmoid-like behaviour mentioned in Sec.~\ref{sec:press-hardening}, and this extends to the monotonized models. An unconstrained polynomial with $m=3$ was chosen as the reference model to be monotonized for the comparative methods from the literature. Degrees lower than that result in larger deviations from the data and degrees higher than that result in an overfit. Thus, out of all models of the form~\eqref{eq:poly-regr-model}, the hyperparameter choice $m=3$ yields the lowest RMSE values for projection and rearrangement. For the monotonic regression with SIAMOR, $m=6$ and five equidistant initial constraint locations in $X^0$ were chosen. It converged in iteration 8 with a total of 12 monotonicity constraints. In terms of the root-mean-squared error, the SIAMOR model predicts the training data more accurately, as can be seen in Tab.~\ref{tab:IWU1D_RMSE}. The reason is that the rearrangement- and projection-based models are dragged away from the data by the underlying reference model, especially at high $T_{\mathrm{f}}$.

\begin{table}
\centering
\caption{Root-mean-squared deviations (RMSE) of the monotonic regression models from the training data for forming and press hardening of sheet metal (1D)}
\label{tab:IWU1D_RMSE}
\begin{tabular*}{0.6\columnwidth}{ll}
\hline\hline
Monotonic regression type & \hspace{3.5em}RMSE [HV] \\
\hline
projection \cite{Lin.2014} & \hspace{3.5em}5.0893 \\
rearrangement \cite{Dette.2006} & \hspace{3.5em}4.8346 \\
SIAMOR & \hspace{3.5em}3.3583\\
\hline\hline
\end{tabular*}
\end{table}

\begin{figure}[htpb]
\centering
\begin{subfigure}[b]{0.45\textwidth}
\centering
\includegraphics[width=\columnwidth]{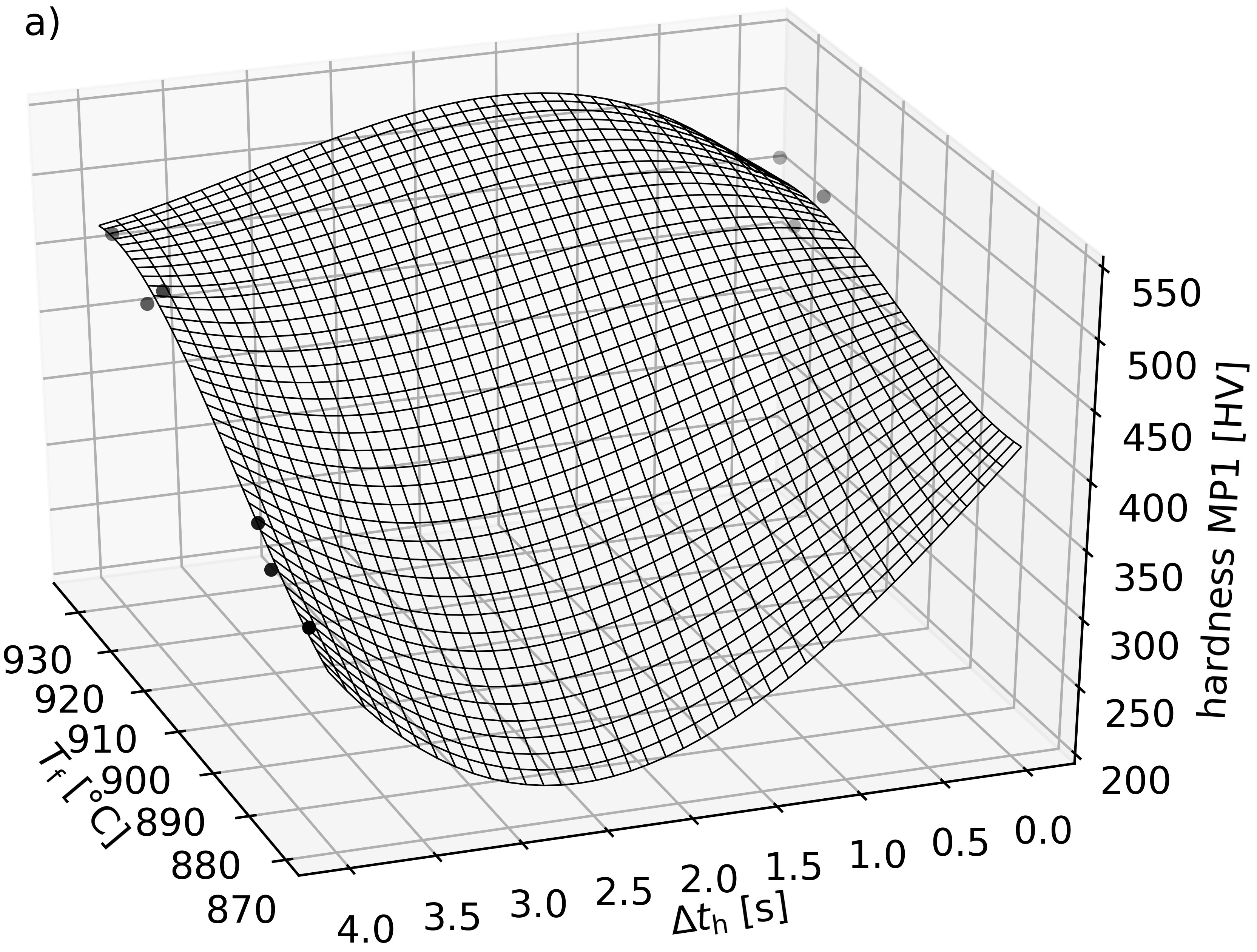}
\end{subfigure}
\hfill
\begin{subfigure}[b]{0.45\textwidth}
\centering
\includegraphics[width=\columnwidth]{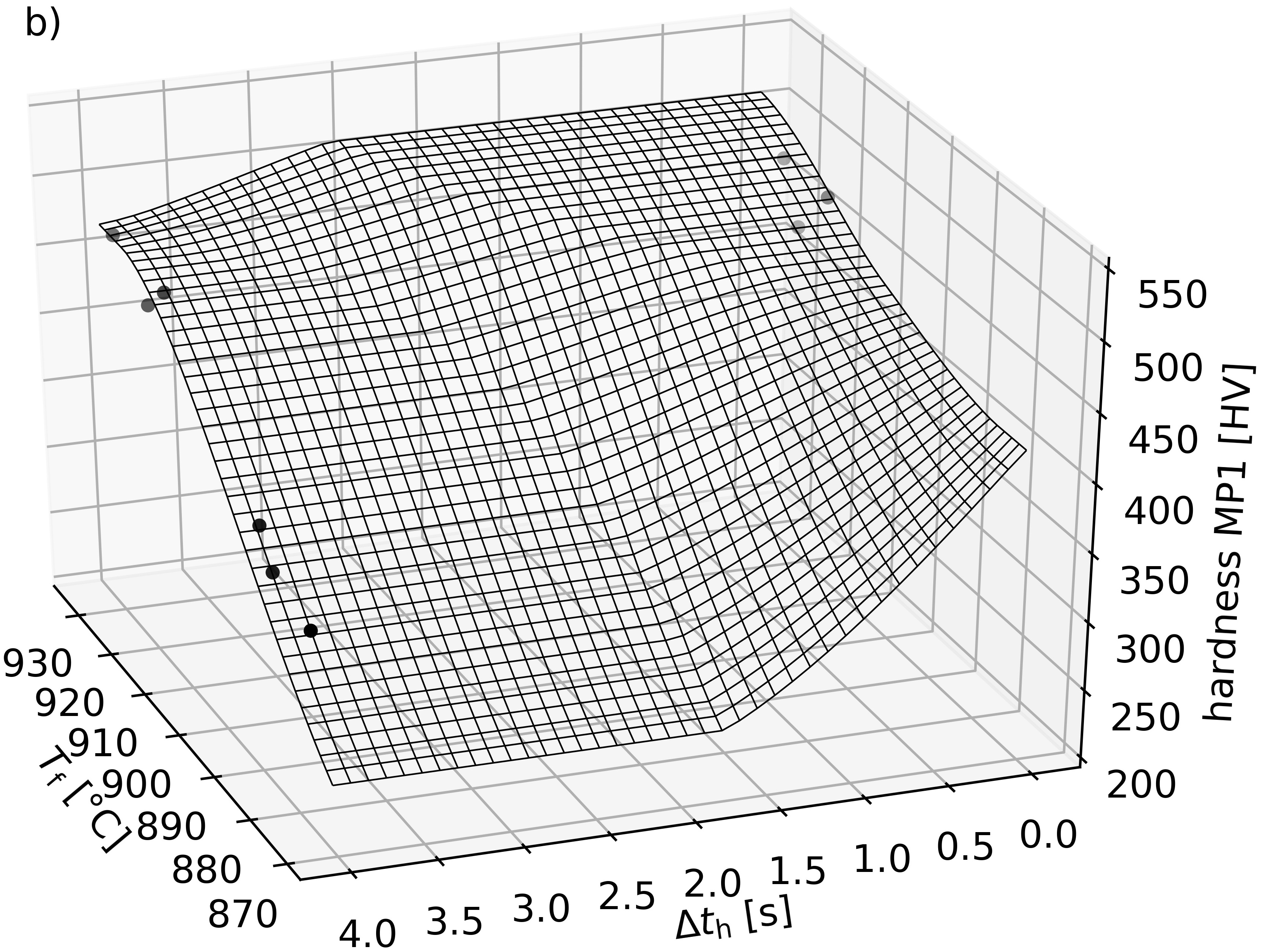}
\end{subfigure}
\vspace{0.5cm}
\begin{subfigure}[b]{0.45\textwidth}
\centering
\includegraphics[width=\columnwidth]{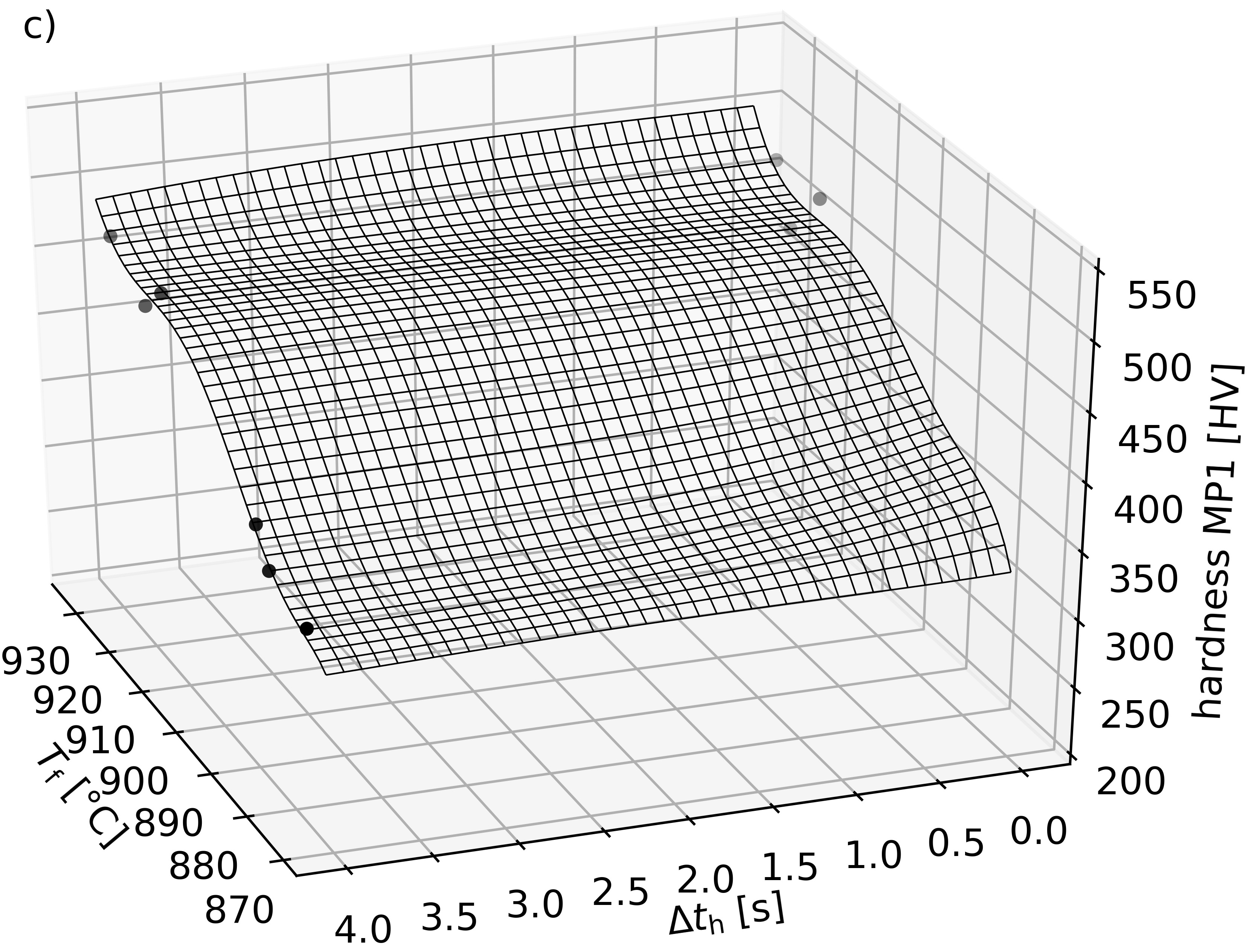}
\end{subfigure}
\caption{4D regression for forming and press hardening of sheet metal using polynomial models ($F_{\mathrm{p}}=2250\,$kN, $t_{\mathrm{q}}=2\,$s). The markers represent those training points matching the specification of the corresponding plane in the input space. a)~Unconstrained $m=3$, b)~projection~\cite{Lin.2014} of unconstrained $m=3$, c)~SIAMOR $m=6$.}
\label{fig:IWU4D}
\end{figure}

After these 1D considerations, the SIAMOR method is now validated on the full 4D data set of the press hardening process. Polynomial models with degree $m=3$ are used for unconstrained regression and monotonic projection, and polynomials with $m=6$ are used for the SIAMOR method. The resulting models are visualized in the surface plots in Fig.~\ref{fig:IWU4D}. The unconstrained model from Fig.~\ref{fig:IWU4D}a clearly shows non-monotonic predictions w.r.t.~$\Delta t_{\mathrm{h}}$. Furthermore, the hardness is slightly decreasing with the furnace temperature at $T_{\mathrm{f}}$ close to 930$\,^\circ$C, which is not the behaviour expected by the process expert either. Fig.~\ref{fig:IWU4D}b shows the monotonic projection of the unconstrained model. It was computed according to Sec.~\ref{sec:computing-mon-proj} on a grid $G$ consisting of 40$^4$ points. The monotonic projection exhibits the kinks that are characteristic of that method and it yields an RMSE of 28.84 HV on the entire data set.

With an overall RMSE of 10.14 HV, the model resulting from SIAMOR is more accurate for this application. A corresponding response surface is displayed in Fig.~\ref{fig:IWU4D}c. In keeping with~\eqref{eq:mon-signature-press-hardening}, monotonicity was required w.r.t.~$T_{\mathrm{f}}$ (increasing), $\Delta t_{\mathrm{h}}$ (decreasing) and $t_{\mathrm{q}}$ (increasing). As $m=6$, there were $N_m=210$ model parameters and the dicretization $X^0$ was initialized with a grid using four equidistant values per dimension. The algorithm converged in iteration 246 with 1372 final constraints. Our first try was with only two monotonicity requirements (namely w.r.t.~$T_{\mathrm{f}}$ and $t_{\mathrm{q}}$) with the observation that the final number of iterations decreases when the third monotonicity requirement is added. Thus, the monotonicity requirements in each direction promote each other numerically within the algorithm and for the used data. Yet, this reduction in the number of iterations is not accompanied by a decrease in total calculation time because more lower-level problems have to be solved when there are more monotonicity directions.

\begin{figure}[htpb]
\centering
\begin{subfigure}[b]{0.45\textwidth}
\centering
\includegraphics[width=\columnwidth]{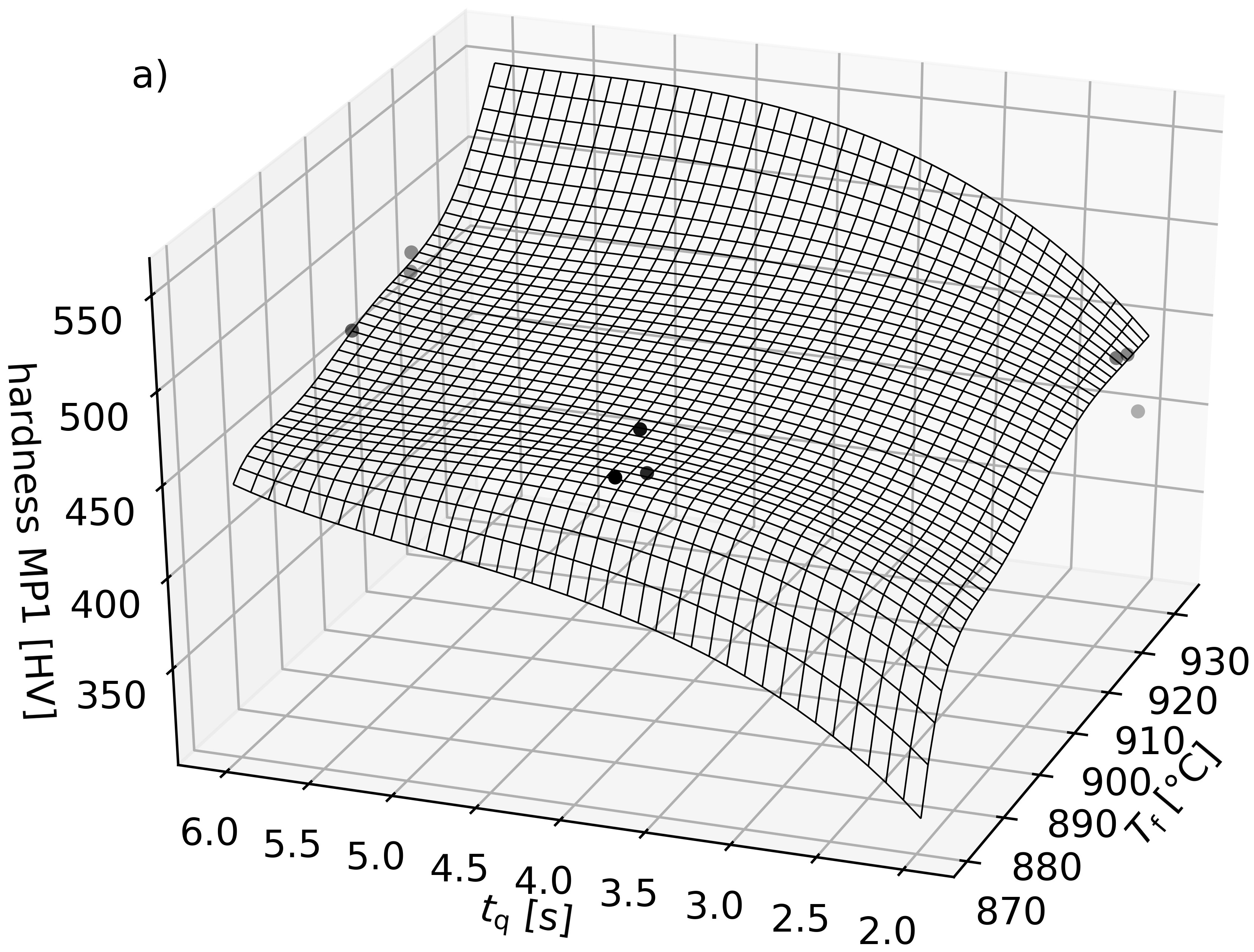}
\end{subfigure}
\hfill
\begin{subfigure}[b]{0.45\textwidth}
\centering
\includegraphics[width=\columnwidth]{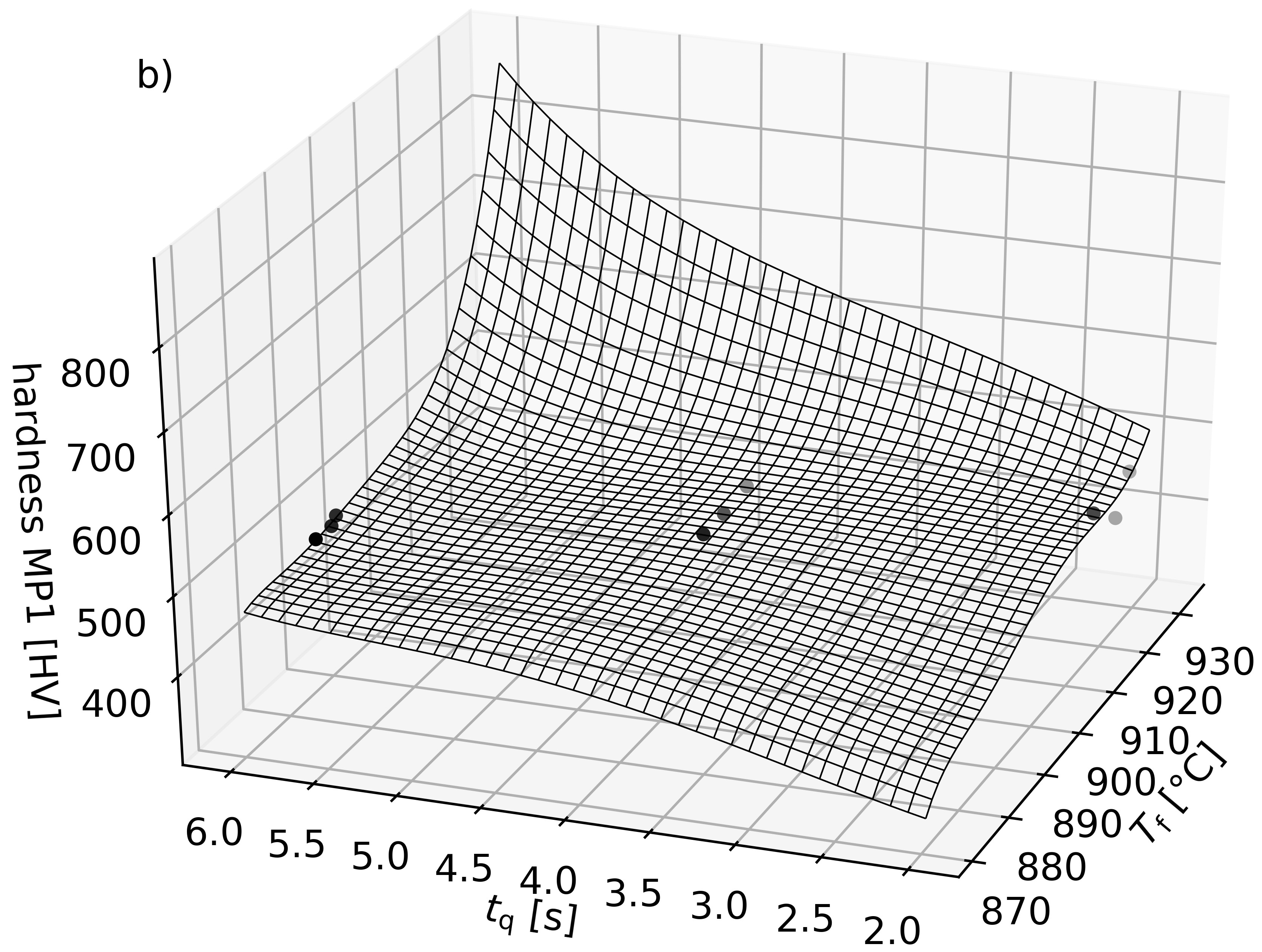}
\end{subfigure}
\caption{Response surfaces of 4D monotonic regression with SIAMOR ($m=6$) for forming and press hardening of sheet metal ($\Delta t_{\mathrm{h}}=0\,$s). The markers represent those training points matching the specification of the corresponding plane in the input space. a)~$F_{\mathrm{p}}=1750\,$kN, b)~$F_{\mathrm{p}}=2250\,$kN.}
\label{fig:IWU4D_2}
\end{figure}

With SIAMOR, monotonicity was achieved in all three input dimensions where it was required. See e.g. Fig.~\ref{fig:IWU4D}c, which is the monotonic counterpart of Fig.~\ref{fig:IWU4D}a. A comparison of Figs.~\ref{fig:IWU4D}a--c clearly shows how incorporating monotonicity expert knowledge helps compensate data shortages. Indeed, taking no monotonicity constraints into account at all (Fig.~\ref{fig:IWU4D}a), one obtains an unexpected hardness minimum w.r.t. $\Delta t_{\mathrm{h}}$ at $\Delta t_{\mathrm{h}} \approx 2.5\,$s and small $T_{\mathrm{f}}$. This also results in unnecessarily low predictions of the monotonic projection for small $T_{\mathrm{f}}$ and $\Delta t_{\mathrm{h}} \gtrapprox 1.5\,$s in Fig.~\ref{fig:IWU4D}b. The SIAMOR model (Fig.~\ref{fig:IWU4D}c), by contrast, predicts more reasonable hardness values in this range without needing additional data because it integrates the available monotonicity knowledge already in the training phase.

For the SIAMOR plots in Fig.~\ref{fig:IWU4D_2}, $\Delta t_{\mathrm{h}}$ was reduced to 0$\,$s. It shows that monotonicity is also achieved w.r.t. $t_{\mathrm{q}}$. Without having explicitly demanded it, the hardness $y$ shows the expected concave growth towards saturation w.r.t. $t_{\mathrm{q}}$ in Fig.~\ref{fig:IWU4D_2}a. An additional increase in $F_{\mathrm{p}}$ leads to Fig.~\ref{fig:IWU4D_2}b, where the sign of the second derivative of $y$ w.r.t.~the quenching time $t_{\mathrm{q}}$ changes along the $T_{\mathrm{f}}$-axis. I.e., the model changes its convexity properties in this direction and increases convexly instead of concavely with $t_{\mathrm{q}}$ at high $T_{\mathrm{f}}$, $\Delta t_{\mathrm{h}}=0\,$s and $F_{\mathrm{p}}=2250\,$kN. This contradicts the process expert's expectations. A possible way out is to measure additional data (e.g.~in the rear left corner of Fig.~\ref{fig:IWU4D_2}b), which is elaborate and costly, however. Another possible way out is to add the concavity requirement $\partial_{x_4}^2\widehat{y}_{\bm{w}}(\bm{x})\le 0$ for all $\bm{x}\in X$ w.r.t.~the $x_4 = t_{\mathrm{q}}$ direction to the monotonicity constraints~\eqref{eq:mon-constraints} used exclusively so far. In order to solve the resulting constrained regression problem, one can use the same adaptive semi-infinite solution strategy which was already used for the monotonicity constraints alone.

\section{Conclusion and outlook} \label{sec:conclusion}

In this article, a proof of concept is conducted for the method of semi-infinite optimization with an adaptive discretization scheme to solve monotonic regression problems (SIAMOR). The method generates continuously differentiable models and its use in multiple dimensions is straightforward. Polynomial models were used, but the method is not restricted to this type of model, even though it is numerically favourable because polynomial models lead to convex quadratic upper-level problems. The monotonic regression technique is validated by means of two real-world applications from manufacturing. It results in predictions that comply very well with expert knowledge and that compensate the lack of data to a certain extent. At least for the small data sets considered here, the resulting models predict the training data more accurately than models based on the well-known projection or rearrangement methods from the literature.

While the present article is confined to regression under monotonicity constraints, semi-infinite optimization can also be exploited to treat other types of shape constraints such as concavity constraints, for instance. In fact, the shape constraints can be quite arbitrary, in principle. And this is only one of several aspects in the field of potential research on the method opened up by this work. Others are the testing of SIAMOR in combination with different model types, data sets or industrial processes. When using Gaussian process regressors instead of the polynomial models employed here, one can try out and compare various kernel types. Additionally, the SIAMOR method can be extended to locally varying monotonicity requirements (i.e. $\sigma_j=\sigma_j(\bm x)$). 

Another possible direction of future research is to systematically investigate how to speed up the solution of the global lower-level problems. When more complex models or shape constraints are used, this becomes particularly important. The solution of multiple lower-level problems and the final feasibility test on the reference grid can be parallelized to reduce the calculation time, for example. A rigorous investigation of the convergence properties and the asymptotic properties of the SIAMOR method and its possible generalizations is left to future research as well.

\section*{Acknowledgements}
This work was supported by the Fraunhofer Society within the lighthouse project `Machine Learning for Production' (ML4P). Our thanks go to Jan Schwientek for his highly valuable advice concerning semi-infinite optimization.

\bibliographystyle{abbrv}
{\small
\bibliography{FraunhoferML4P_SIAMOR}
}

\end{document}